\documentclass[11pt]{article}

\usepackage{amsmath}
\usepackage{amssymb}
\usepackage{graphicx}
\usepackage{hyperref} 
\usepackage{geometry} 
\usepackage{doi}      
\usepackage{adjustbox}
\usepackage{pdflscape}
\usepackage{multirow}
\usepackage{url}
\usepackage{subcaption}

\usepackage[russian, english]{babel}

\usepackage{lscape}

\usepackage{bm}

\title{Deep Learning Models Meet Financial Data Modalities}
\author{
  Kasymkhan Khubiyev \\
  Sirius University of Science and Technology, Sirius, Russia \\
  \texttt{kasymkhankhubievnis@gmail.com}
\and
  Mikhail Semenov\\
  Sirius University of Science and Technology, Sirius, Russia \\
  \texttt{semenov.me@talantiuspeh.ru}
}
\date{\today}

\begin{document}

\maketitle

\footnotetext{Preprint for the \textit{MathAI: Mathematics of Artificial Intelligence} conference
}

\begin{abstract}
Algorithmic trading relies on extracting meaningful signals from diverse financial data sources, including candlestick charts, order statistics on put and canceled orders, traded volume data, limit order books, and news flow. While deep learning has demonstrated remarkable success in processing unstructured data and has significantly advanced natural language processing, its application to structured financial data remains an ongoing challenge. This study investigates the integration of deep learning models with financial data modalities, aiming to enhance predictive performance in trading strategies and portfolio optimization.

We present a novel approach to incorporating limit order book analysis into algorithmic trading by developing embedding techniques and treating sequential limit order book snapshots as distinct input channels in an image-based representation.

Our methodology for processing limit order book data achieves state-of-the-art performance in high-frequency trading algorithms, underscoring the effectiveness of deep learning in financial applications.
\end{abstract}

{\centering
\section{Introduction}
}
Algorithmic trading is a domain that continually seeks new solutions and requires the researcher to extract new signals from the data flow to keep the trading results best and increase the profit. Mainly strategies are built by a scrutiny analysis of existing data in an attempt to reveal latent insights from trading data in its wide context: historical candlestick time series, order books, traded volumes statistics, annual reports, etc. Due to tremendous success of deep learning (DL) in various fields processing image, audio and text, it is essential to fit deep learning technologies to financial use cases. 

Besides common understanding of data modalities: numeric, text, audio, image, there are finance-specific modalities in trading data flow, for example, a limit order book (LOB). Separating data based on their nature is essential for the design of trading strategies. LOB data are commonly used in  high frequency trading (HFT) to extract current market state and predict market continuous dynamic to perform efficient market-making. 

Authors~\cite{Ntakaris2017BenchmarkDF} propose a banchmark dataset for mid-price forecasting of LOB data. They scrapped 4 million LOB snapshots dataset from NASDAQ Nordic stock market and used ML algorithms to forecast LOB mid-price. They applied z-score and decimal precision normalization, min-max scaling for 10 levels depth LOB data, where the \textit{depth} of a LOB is the amount of unique prices (levels) of sell or buy prices. They performed multiclass classification task, predicting labels that indicates the percentage change of LOB mid-price data. The authors~\cite{Ye2024} states two streams of HFT researches: model-based and data-driven. The first stream focuses on methods and models constructed on price dynamics stochastic properties. The second stream focuses on data constructing models and method based on empirical observations to predict and design properties. Follow the second stream, the authors proposed a novel imaging framework that converts high-frequency LOB data into standardized 2-dimensional grayscale images to predict the mid-price movement in the next 30 seconds. Their datasets contained LOB data of 20 stocks listed on Shanghai Stock Exchange and included into CSI 300 index. The LOB snapshot embedding technique took a crucial role in the research, the introduced approach enhanced the efficiency and performance of short-term stock price trend prediction using convolutional neural networks (CNNs).

Authors~\cite{BRIOLA2025126078} address the issue of LOB mid-price trend forecasting. They analyzed 15 stocks traded 
on the NASDAQ exchange, 
having them divided into three groups based on their average bid-ask spread: small-, medium- and large-tick stocks. The target value is the LOB mid-price trend state: up, stable, down. The authors observed 10, 50 and 100 ticks forecasting horizon, having the horizon defined in terms of LOB states updates, not real time. They introduced HLOB model -- a large scale DL model with Triangulated Maximally Filtered Graph (TMFG) in a core, three CNN blocks to process tetrahedra, triangles and edges, and LSTM with MLP layers. They removed price columns from LOB data, volume is categorized into discrete bins to build Information Filtering Network (IFN). They showed that the proposed HLOB model better catches dependencies and intersections between consecutive LOB levels leveraging the TMFG.

The authors~\cite{AngererLOBCrypto2025} scrutinized cryptocurrencies market across several exchanges to reveal various trade essential patterns from LOB data. They highlighted that most studies of LOB liquidity focus on the best bid-ask levels volumes and prices, but the information collected in these levels are often noisy, while deeper LOB levels often provide more informative profile of the current liquidity. They claim that the orders concentration in a LOB reflects a market state, short-term volatility and market participants expectations. For example, a high volume concentration around best quotes indicates a market consensus on the asset true price and lower volatility, while the volume dispersed over depth indicates volatility increase. The authors notice that it is important to investigate the LOB resilience -- the ability to recover from orders flow anomalies, to forecast possible crisis events and design more robust high frequency strategies. For the research, the authors used dataset of 11 unique base cryptocurrencies including BTC and ETH and trading pairs from four cryptoexchanges: OKeX, Binance, Huobi, and Kraken. LOBs depth varied between exchanges, for example, while OkeX provided full depth, Binance provided only first 20 levels for both ask and bid sides. To perform an analysis the authors filtered data dropping LOB snapshots with data gaps and transformed data points into 5 minutes frequent 20 levels depth LOB snapshots. They concluded that bid- and ask-sides depth comparative description plays a crucial role in market stability evaluation, underscoring spread, depth and volume imbalance as an important measure of liquidity and a crucial base for decision making. 

In the current research we investigate LOB data, embedding techniques and mid-price forecasting to design trading strategies. LOB data is commonly used in the high frequency trading (HFT) to extract current market state and predict market continuous dynamic to perform efficient market-making. We treat LOB data as a unique data modality and highlight methodologies to benefit from it.

The key contributions of this paper are as follows.
\begin{enumerate}
    \item We introduce LOB data as an ``in-finance`` modality in essence of the data nature and treat and propose several LOB data embedding techniques.
    \item We design high frequency trading strategy based on LOB data forecasting with cryptocurrencies.
\end{enumerate}

The paper is organized as follows. 
In Section 2, we briefly describe the original dataset and conduct an exploration of it. 
In Section 3, we present a research methodology that covers evaluation metrics, custom loss functions, limit order book sampling, and embedding techniques. 
In Section 4, we describe performed experiments set up and pipelines, introduce models architectures and algorithmic trading strategies.
In Section 5, we describe the results of computational experiments.
Finally, Section~6 concludes the paper and discusses future work.


{\centering
\section{Data}
}

In the current paper, we examine a novel approach handling a limit order book (LOB) data as a \textit{specific financial data modality}. In contradiction to historical candlesticks data -- a common time-series entity -- LOB data represent a snapshot of a LOB instant state at the specific time. Based on the nature of LOB and simple heuristics, the idea of implementing well-known and efficient computer vision techniques naturally fits the task of mid-price forecasting. Although financial data are usually well structured, the LOB state is chaotic and suitable for DL models that are proven to be the best in dealing with unstructured data.

To examine LOB data-based trading strategies in various settings, we needed high frequency LOB data. We collected 
Bitcoin (BTC), 
Etherium (ETH), 
Solana (SOL), 
Cardano (ADA), 
Tron (TRX), 
Telegram Open Network (TON), 
Binance coin (BNB), 
DOGEcoin (DOGE), and 
GOATcoin (GOAT) 
perpetual futures limit order books snapshots from November 27, 
2024 to November 29, 2024 from the cryptocurrency exchange ByBit to perform experiments. The time difference between two sequential LOB snapshots is approximately 200-300~ms.

Each LOB snapshot is described by the set of \textit{ask} prices (orders to sell) and \textit{bid} prices (orders to buy). Ask prices are always higher than bid prices. The difference between the minimum ask price and maximum bid price is called \textit{spread}. The minimum ask price is called \textit{best ask}, $p_a$ and the maximum bid price is called \textit{best bid}, $p_b$. The \textit{mid-price}, $p_{mid}$ is calculated as follows:
\begin{equation}
    p_{mid} = \frac{1}{2}(p_a + p_b).
\label{mid_price_eq}
\end{equation}
The amount of unique prices (levels) of ask or bid prices is called the \textit{depth} of a LOB. The collected data consists of LOB snapshots of the depth of $D=50$, i.\,e. there are 50 unique prices for both ask and bid. Based on orders execution attitude are divided into market and limit orders. If the order executes at the same time it was put and executed with best available price is called \textit{market} order and the order owner is called \textit{taker}. Otherwise the \textit{limit} order is put with intent to be executed later providing liquidity. The owner of the executed limit order is called \textit{maker}.

{\centering
\section{Methodology}
}
We examined different approaches of LOB data processing in perspective of mid-price forecasting and algorithmic trading strategies (alphas)~\cite{Kakushadze2016}.  We propose a novel approach with DL models and build high frequency trading (HFT) strategy. We evaluate the models inference with mean absolute percentage error (MAPE) of mid-price forecast, Sharpe ration and Profit values of a trading strategy.

{\centering\subsection{Sampling and Embedding Techniques}}

In HFT and automated market-making (AMM) mid-price forecasting is essential. We treat the LOB data as a discrete sequence of LOB instant state snapshots and provide mid-price or its increment sign as a target value. Having a sequence of $N$ $2$-dimensional snapshots at time~$t$ we predict the mid-price value or trend direction for a time $\tau = t + \Delta t$, having $\Delta t$ equal to 1~second.

The central issue is data embedding. Let us 
assume a set of ask prices sorted in ascending order be 
$\mathbf{a} = \left(a_{[1]}, a_{[2]}, \ldots, a_{[D]}\right)$ 
and a set of bid prices sorted in descending order be 
$\mathbf{b} = \left(b_{[1]}, b_{[2]}, \ldots, b_{[D]}\right)$. We implement asks and bids prices scaling regarding to a specific LOB mid-price 
accordingly:
\begin{equation}
\tilde{a}_{[i]} = \frac{a_{[i]} - p_{mid}}{\max(\mathbf{a}) - p_{mid}}, 
\quad
\tilde{b}_{[i]} = \frac{b_{[i]} - p_{mid}}{\min(\mathbf{b}) - p_{mid}},
\quad 
i=1, 2, \ldots, D.
\label{a_price_bp_scale}
\end{equation}
The ask and bid prices scaling functions embodies the idea of \textit{min-max scaling} regarding to mid-price. The idea is to transform prices into interval from 0 to 1 dividing by the initial interval length, for ask prices the interval length is $(\max(\mathbf{a}) - p_{mid})$ and for bid prices the length is
$(\min(\mathbf{b}) - p_{mid})$.

To adders orders value scaling issue we examined three types of scaling: 
z-score scaling, 
\textit{min-max} like scaling over global values 
and domain specific values 
respectively: 
\begin{equation}
\tilde{q}_i = \frac{q_i - \mathbb{E}(q)}{\sigma(q)},
\label{z_score_scale}
\end{equation}
\begin{equation}
    \tilde{q}_i = \frac{q_i - MIN_{global}}{MAX_{global}-MIN_{global}}
    \label{quantity_global_scaling},
\end{equation}
\begin{equation}
    \tilde{q}_i = \frac{q_i - MIN_{domain}}{MAX_{domain}-MIN_{domain}}, 
    \quad domain \in \{ask, bid\}
    \label{quantity_domain_skaling},
\end{equation}
where $\mathbb{E}(q)$, $\sigma(q)$ -- 
expected value and standard deviation, $MIN_{doman}$ and $MAX_{domain}$ are minimum and maximum value over subsample and $MIN_{global}$ and $MAX_{global}$ are minimum and maximum value over the whole sample of random variable $q$ respectively.
Speaking about \textit{domain} we consider minimum and maximum order quantities over ask or bid orders.

\begin{figure}[ht!]
    \centering

    \begin{subfigure}[t]{0.3\textwidth}
        \centering
        \includegraphics[width=\textwidth]{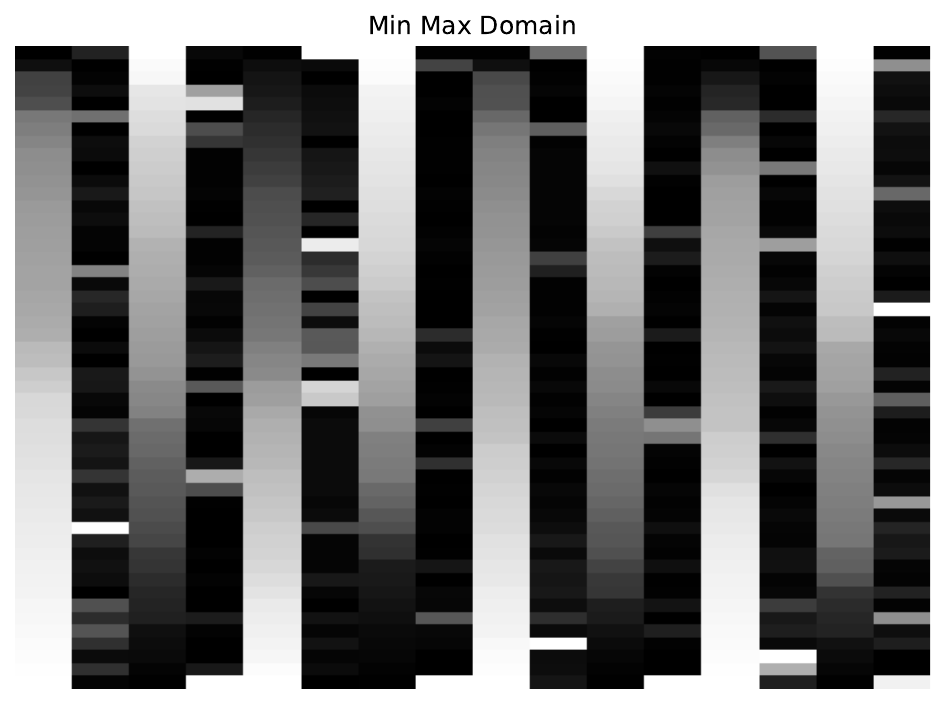}
        \caption{}
        \label{fig:sub1}
    \end{subfigure}
    \hfill %
%
    \begin{subfigure}[t]{0.3\textwidth} 
        \centering
        \includegraphics[width=\textwidth]{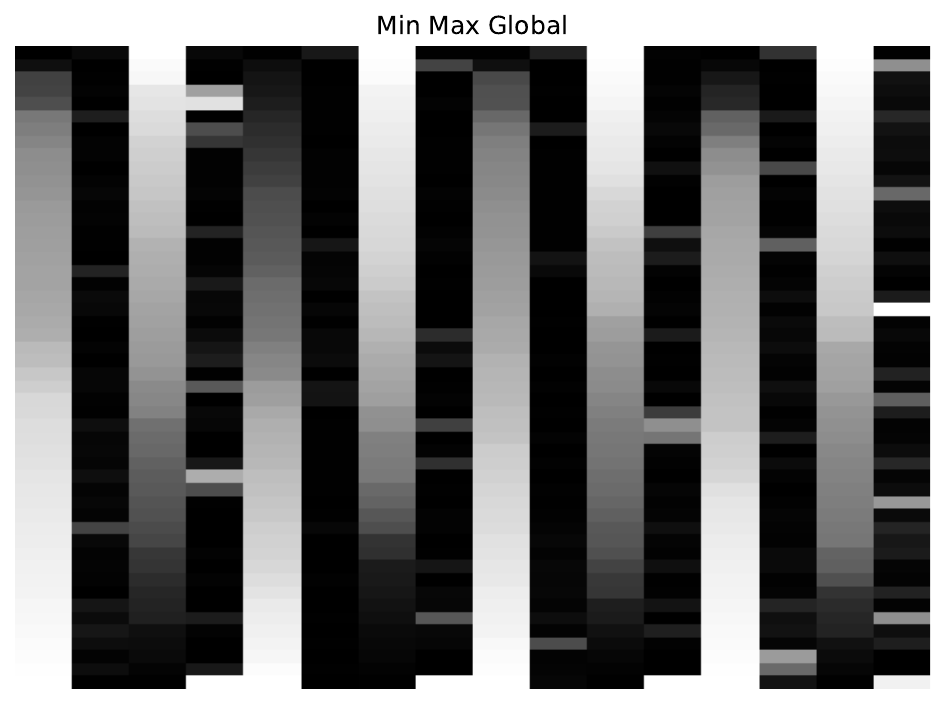} 
        \caption{}
        \label{fig:sub2}
    \end{subfigure}
    \hfill 
%
    \begin{subfigure}[t]{0.3\textwidth} 
        \centering
        \includegraphics[width=\textwidth]{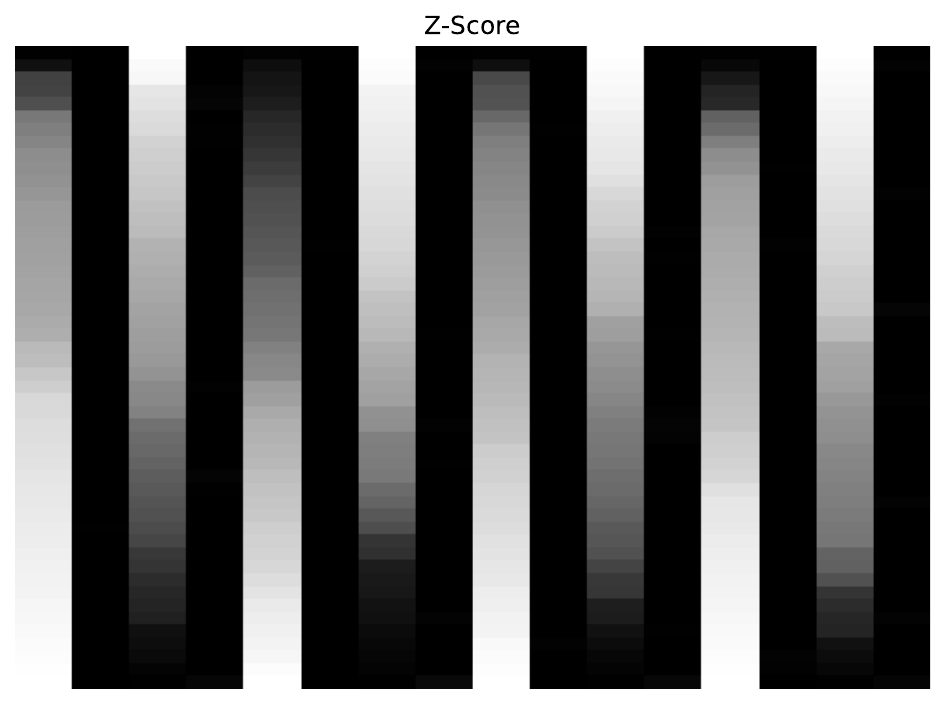} 
        \caption{}
        \label{fig:sub3}
    \end{subfigure}

    \caption{Types of scaling: (a) min-max domain, (b) min-max global, (c) z-score}
    \label{fig:overall}
\end{figure}

These prices scaling technique is essential because a current LOB state mid-price is incorporated into the scaling factor. After scaling~\eqref{z_score_scale}-\eqref{quantity_domain_skaling} we multiply values by 255, so the final matrices values lie between 0 and 255, making the input values sets similar to images channels input. The orders value scaling must best represent its distribution over the LOB in order to 
the DL model could benefit from it. Having a closer look at LOB snapshots transformed and merged into a single 2-dimensional image (Figure~\ref{fig:overall}), we see that z-score scaling does not express any information over orders quantity distribution, while min-max domain scaling offers the most informative view of the data. Considering that in HFT we send \textit{sell} and \textit{buy} orders simultaneously, we chose min-max domain scaling. 

\begin{figure}[ht!]
    \centering

    \begin{subfigure}[t]{0.4\textwidth}
        \centering
        \includegraphics[width=\textwidth]{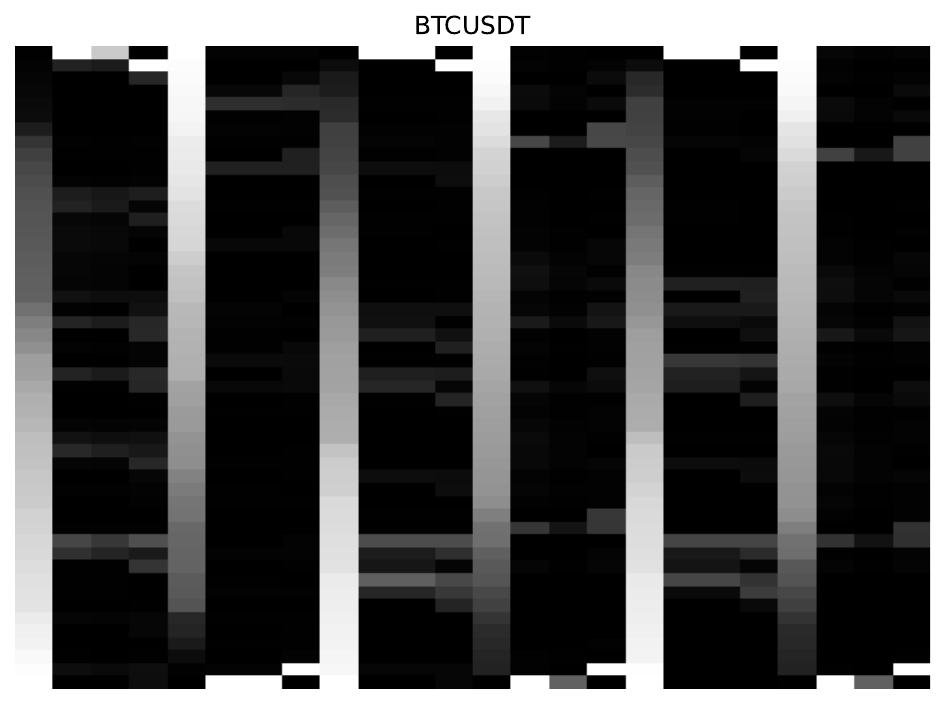}
        \caption{}
        \label{fig:btcusdt}
    \end{subfigure}
    \hfill %
%
    \begin{subfigure}[t]{0.4\textwidth} 
        \centering
        \includegraphics[width=\textwidth]{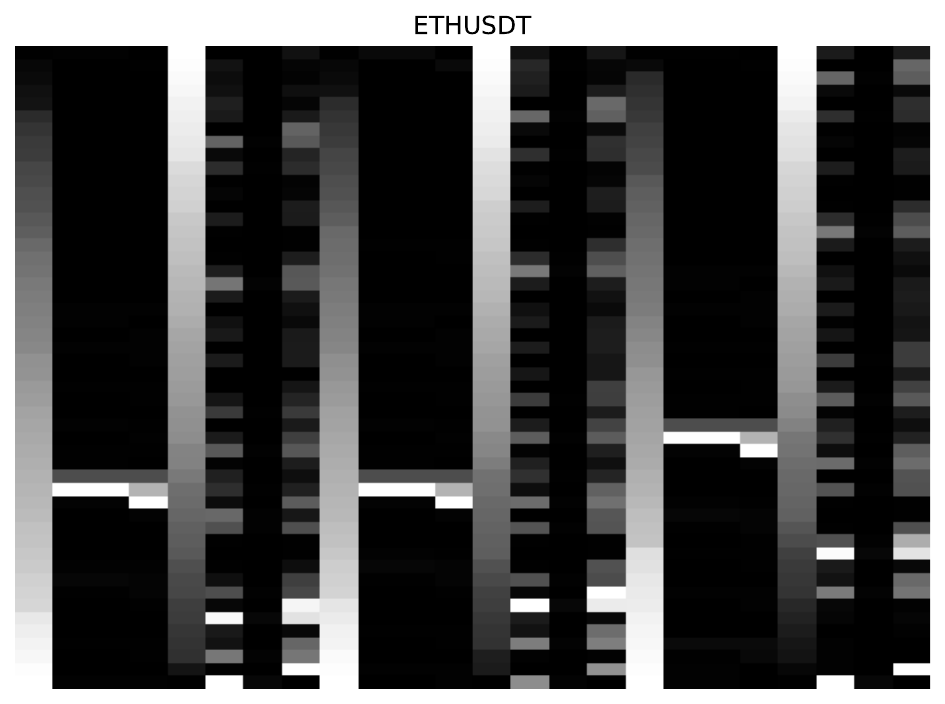} 
        \caption{}
        \label{fig:ethusdt}
    \end{subfigure}
    \hfill
%
    \begin{subfigure}[t]{0.4\textwidth} 
        \includegraphics[width=\textwidth]{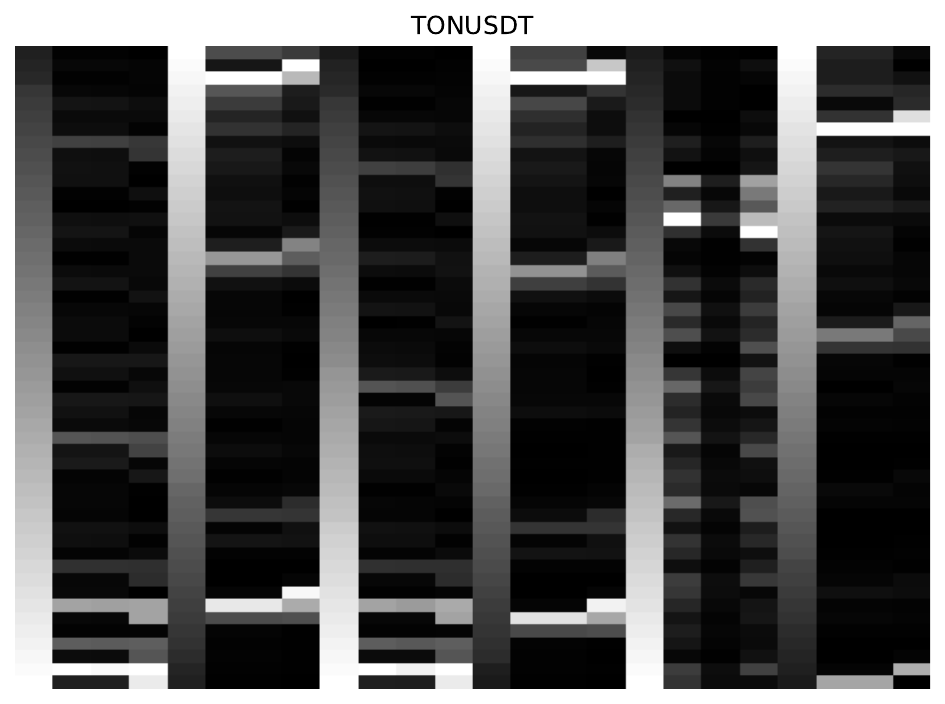} 
        \caption{}
        \label{fig:tonusdt}
    \end{subfigure}
    \hfill
%
    \begin{subfigure}[t]{0.4\textwidth} 
        \centering
        \includegraphics[width=\textwidth]{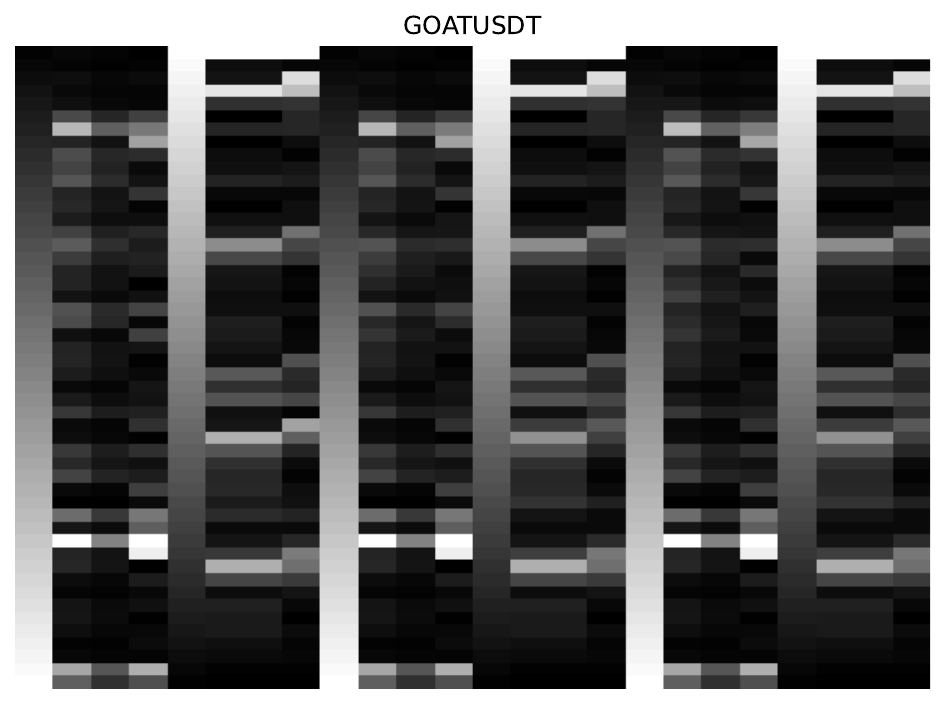} 
        \caption{}
        \label{fig:goatusdt}
    \end{subfigure}
    \hfill

    \caption{Transformed LOB snapshots of different derivatives: 
    (a) BTC, (b) ETH, (c) TON, (d) GOAT}
    \label{fig:coins_lobs}
\end{figure}

Following~\cite{AngererLOBCrypto2025} we wanted to incorporate information about volume imbalance and current price and volume bins. For example, using only \textit{Min-Max-Domain} scaling lacks cross-domain information i.\,e. current imbalance state. To address the issue, we added complementary features: \textit{min-max-global} scaling column and consecutive volume and price levels bins width. Incorporating information about levels bins width is essential to capture market changes from sequential LOB snapshots. For example, if the bins width expands it indicates liquidity decrease because empty levels exist, contrariwise narrowing bins indicate liquidity influx. Figure~\ref{fig:coins_lobs} demonstrates examples for transformed LOB snapshots: 
BTC, ETH, TON, GOAT.

{\centering\subsection{Sampling Techniques}}

Having solved the scaling issue, we have to determine the optimal data aggregation technique. In the current research, we specify two separate aggregation problems: (a) data representation and (b) historical interval length. 

Historical interval length addresses the problem of choosing the correct time interval in which we have to aggregate LOB snapshots. We offer different approaches. The first approach aggregates LOB snapshots within a rounded time interval, for example, 1 second or 1 minute. It creates \textit{non-overlapping} sequences grouping LOB states within a given time interval including borders as input and the following LOB state mid-price as a target value, then it strides over collected snapshots and gathers another sample. The second approach aggregates a sequential LOB snapshots followed one by another. The method generates \textit{overlapping} sequences, rolling with a size window over LOB snapshots gathering them into an input, having the next sequential LOB mid-price as a target value.

The data representation issue is related to the choice of data structure. For example, Ye\,et.\,al.~\cite{Ye2024} gather LOB snapshots into a single 2-dimensional image. Following Ye et. al.~\cite{Ye2024}, we merge sequential LOB snapshots into a single 2-dimensional image of the shape 
$D\times(4L)$, 
where $L$ -- is the number of snapshots included into the observation, that is why we call the aggregation method \textit{"merged"}. The second approach treats LOB snapshots as image channels. In comparison with a common RGB image, which contains three input channels, the input data of a shape 
$L\times D\times4$ -- has $L$ channels, having each LOB snapshot to be a separate channel, that is why we call it \textit{"stacked"} method.

The problem of optimal prediction horizon remains essential. Having historical data tine interval to be too short might make the model focused on local changes missing long term trends knowledge. Otherwise having historical interval to be too long might affect the model in the way the earliest occasional events bias the model even if they are no longer actual. For the experiment, we choose the following aggregation intervals:
$$
1 s, 2 s, 3 s, 5 s, 10 s; \quad 2 w, 3 w, 5 w, 10 w, 30 w, 
$$
where the postfix \textit{s} and \textit{w} denotes second and window respectively. The process is the following: for $5s$ the dataset is grouped by LOB snapshots observation within 5 seconds, for $5w$ the sequential snapshots are gathered with a sliding window of the length of 5.

To address the prediction horizon issue, we sampled data with the following horizons:

$$
0.2 s, 1 s, 5s, 10s, 30s, 60s,
$$
where the postfix \textit{s} denotes second. The minimal horizon $0.2$~seconds is the frequency of a LOB update, the upped bound is 60~seconds, because in the current research we consider a high frequency case. 

Another important component is a target value -- the value a model predicts. The initial idea behind is to forecast a mid-price. We consider two different types of target values -- price delta 
\begin{equation}
    \Delta p(t) = price(t+1) - price(t) 
    \label{eq:price_delta}
\end{equation}
and price relative growth
\begin{equation}
    r(t) = \frac{price(t+1)}{price(t)} - 1, 
\label{eq:price_returns}
\end{equation} 
where $t\ge 0$ is time point.
If a model predicts the price delta~\eqref{eq:price_delta}, the final result is the last LOB snapshot mid-price plus the predicted delta. If the model predicts the price relative growth~\eqref{eq:price_returns}, there are two ways to transform the predicted return into an appropriate mid-price: (a)~use the last or (b)~use the first LOB snapshot mid-price of the observed sequence.

For the training, we used LOB snapshots dated by November 27, 2024 for each interval having $5,000$ training samples.

{\centering
\section{Experiments}
}

{\centering
\subsection{Models}
}

\begin{figure}[ht!]
    \centering
%
    \begin{subfigure}[t]{0.49\textwidth}
        \centering
        \includegraphics[width=\textwidth]{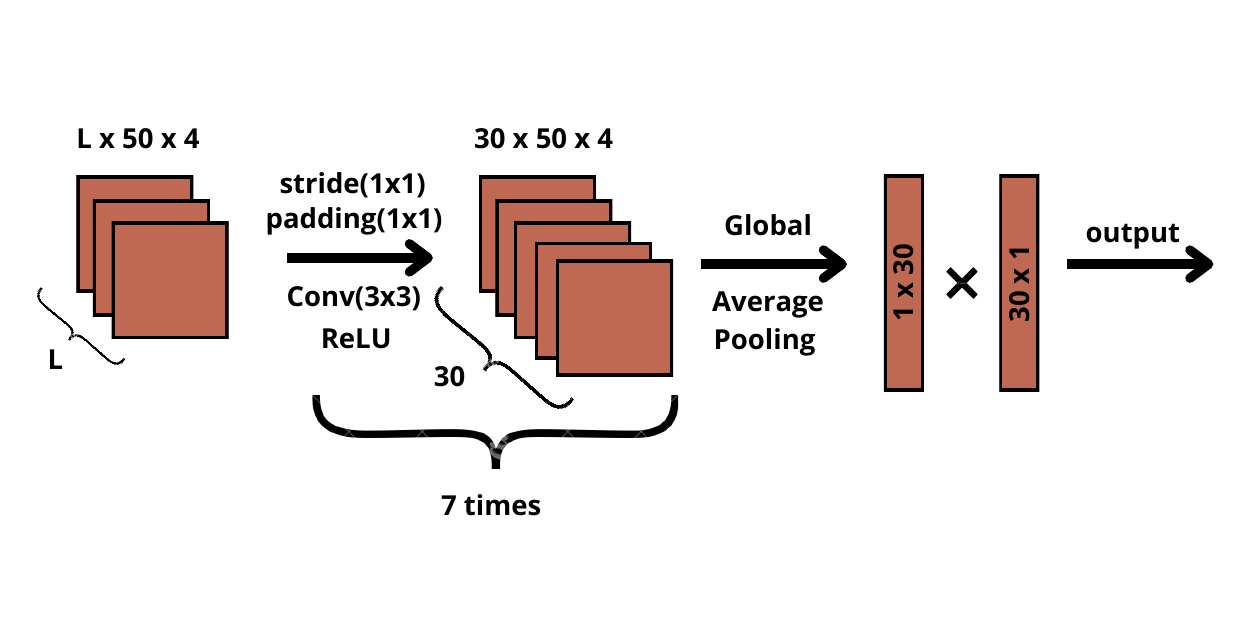}
        \caption{}
        \label{fig:sub_1}
    \end{subfigure}
    \hfill %
 %
    \begin{subfigure}[t]{0.5\textwidth} 
        \centering
        \includegraphics[width=\textwidth]{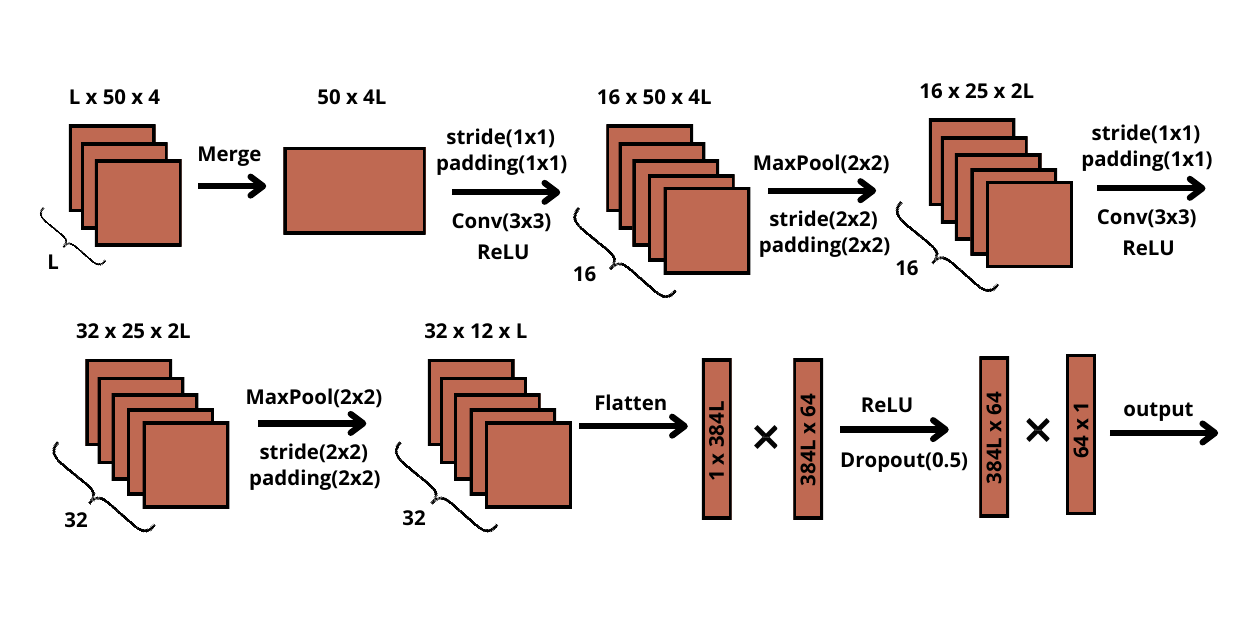} 
        \caption{}
        \label{fig:sub_2}
    \end{subfigure}
    \hfill 
%
    \begin{subfigure}[t]{0.49\textwidth} 
        \centering
        \includegraphics[width=\textwidth]{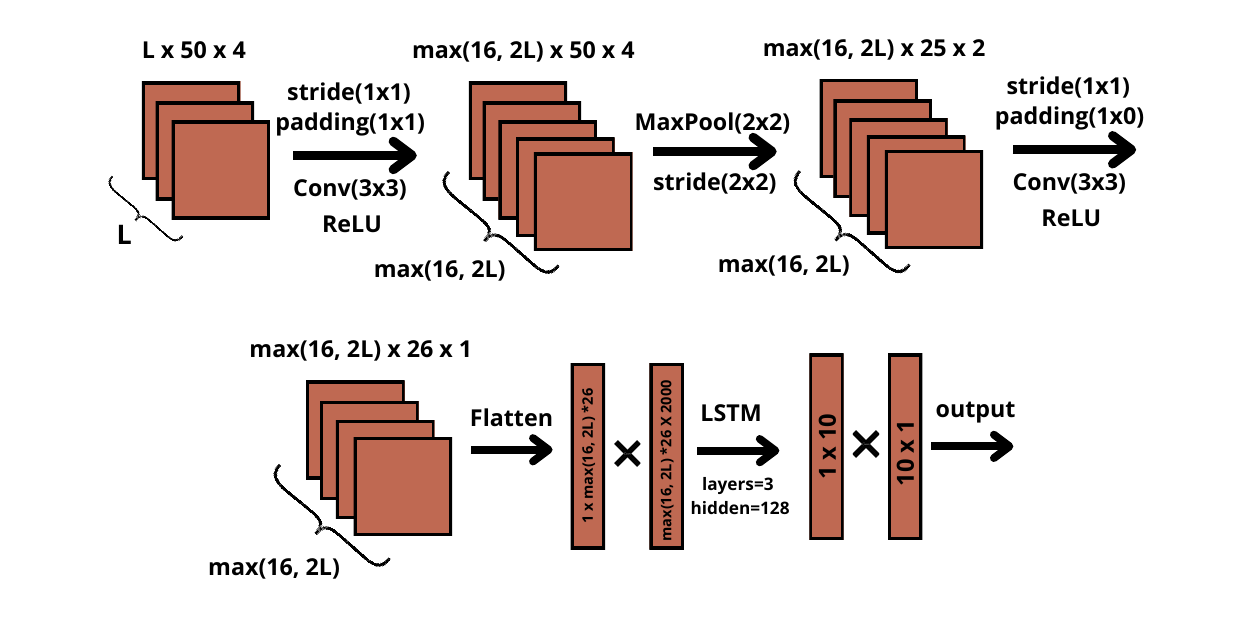} 
        \caption{}
        \label{fig:sub_3}
    \end{subfigure}
    \hfill 
%
    \begin{subfigure}[t]{0.5\textwidth} 
        \centering
        \includegraphics[width=\textwidth]{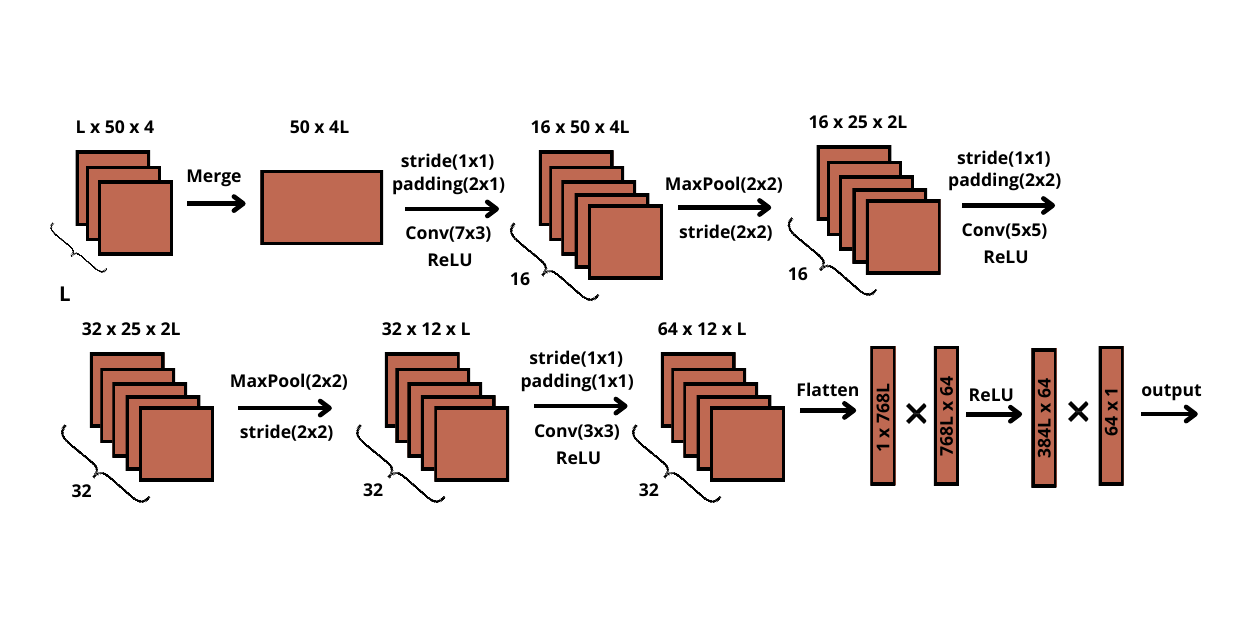} 
        \caption{}
        \label{fig:sub_4}
    \end{subfigure}

    \caption{Models architectures: (a) SimpleCNN, (b) SimpleCNN\_2D,  (c) CNN2LSTM, (d) CNNModel\_2D}
    \label{fig:models}
\end{figure}

To perform the experiment with LOB data, we have designed four models based on convolution layers: 
SimpleCNN (Figure~\ref{fig:sub_1}) -- simple CNN model;  
SimpleCNN\_2D (Figure~\ref{fig:sub_2}) -- simple CNN model with four CNN blocks with MaxPool and MLP for \textit{merged} sequence;
CNN2LSTM (Figure~~\ref{fig:sub_3}) -- CNN model in connection with LSTM model for \textit{stacked} sequence, having CNN model as an embedding layer transforming a sequence of LOB frames into a single vector; and 
CNNModel\_2D (Figure~\ref{fig:sub_4}) -- CNN model model with five CNN blocks with MaxPool and MLP for \textit{merged} sequence.
The CNN2LSTM model converts a sequence of LOB snapshots of a shape $L \times 50 \times 4$ into a vector of length $26 \times \max(16, 2L)$. Here we use a simple technique to expend dimensionality multiplying the CNN output vector with an embedding vector of length $2,000$ to feed it into a LSTM model.

{\centering
\subsection{Trading Strategy}
}

To examine the forecast quality, we performed a trading strategy for automated market making (AMM). Some cryptocurrency exchanges offer negative fees for AMM market makers (MM) trading with weekly or monthly futures. For example, Deribit Exchange offers $-0.01\%$ (1~bips) for MM trading with BTC and ETH futures, and $0.05\%$ (5 bips) for takers. Our strategy mainly focuses on earning with negative fees and having AMM execution costs close to zero.

To perform the AMM trading strategy, we use sliding window with a size of 30 aggregation method for stacked and merged LOB snapshots. We simultaneously placed sell and buy limit orders based on the predicted mid-price considering a spread equal to $0.1$~USD. We limit the order size by 5~USD, so the short and long position are not more than 5~USD each. If the sell order price is greater than best bid or buy order price is less than best ask price at the next moment of the LOB snapshot -- the orders will be executed as market orders; otherwise we compare the orders prices and quantity with the trades history to know whether an order could be executed or not. The main limitation is that we expect the order to be executed for sure because the order size is the minimum available, and we do not take into account the orders queue inside the LOB. To perform the experiments, we used LOB snapshots dated by November 28, 2024, ensuring that model training and test data do not overlap.

{\centering
\subsection{Evaluation Metrics}
}
To evaluate the models mid-price forecasting quality, we used mean absolute percentage error (MAPE): 
\[
MAPE = \frac{\mid p - \hat{p}\mid}{p} \cdot 100\%, 
\] where $p$ is an actual value and $\hat{p}$ -- predicted value. 

To compare different alpha execution results we use the following metrics -- Sharpe ratio~\eqref{eq:Sharpe_ration}, profit and loss (PnL)~\eqref{eq:pnl}, and
Max Drawdown~\eqref{eq:maxdrowdown}:

\begin{equation}
    \text{Sharpe ratio} = \sqrt{N}\frac{\mathbb{E}(\mathbf{pnl})}{\sigma(\mathbf{pnl})}, 
\label{eq:Sharpe_ration}
\end{equation}

\begin{equation}
    \text{PnL} = 
    \sum_{i=1}^N pnl_i, 
\label{eq:pnl}
\end{equation} where $pnl_i$ is a strategy earning in time $i$, $\mathbf{pnl} = (pnl_1, pnl_2, \ldots, pnl_N)$.

\begin{equation}
\text{Max DrawDown} = \min(cumsum(\mathbf{pnl}) - cummax(\mathbf{pnl})),
\label{eq:maxdrowdown}
\end{equation} where \textit{cummax}() and \textit{cumsum}() are cumulative maximum and cumulative sum of a given profit
and loss vector correspondingly.

Sharpe ratio represents how constantly a give alpha earns, the greater the Sharpe ratio value, the more smooth a cumulative PnL curve is and the more constantly the strategy earns. High Sharpe ratio values do not mean huge profit, but mean less states, when the alpha losses money. Max Drawdown indicates the maximum money loss of the strategy, while PnL shows the final profit compared to initial bank account. These metrics are correlated. For example, having large drawdown values affects lower Sharpe ratio.

{\centering
\section{Results and Discussion}}

{\centering
\subsection{Optimal Parameters}}

From the results Table~\ref{tab:table1} one can see that the longer the historical interval is observed, the less \textit{MAPE} value is for all stacked and merged LOBs frames. Interestingly the straight forward method -- predicting the price delta -- performed results with the lowest \textit{MAPE} value. Also forecast having LOB snapshots stacked in average perform better than merged (Table~\ref{tab:table3}). The obvious trend exists reducing the LOB history length degrades the results, but the model with LSTM layer showed in Table~\ref{tab:table2} loss decrease for each sequence length, highlighting its forecasting capability.

\begin{table}[ht!]
\centering
\begin{tabular}{|l|l|l|l|l|}
\hline
\textbf{Sampling} & \textbf{MAPE, \%} & \textbf{Aggregation} & \textbf{Model}       & \textbf{Task}        \\\hline
30w & 0.001511 & stacked & CNN2LSTM & regression \\\hline
30w & 0.001531  & merged & SimpleCNN\_2D & regression \\\hline
10w & 0.001648 & stacked & CNN2LSTM & regression \\\hline
10w & 0.002389 & stacked & SimpleCNN & regression \\\hline
1s & 0.003733 & merged & SimpleCNN\_2D & returns \\\hline
30w & 0.008369 & stacked & CNN2LSTM & returns\\\hline
3w  & 0.011200 & stacked & CNN2LSTM & returns \\\hline
5s  & 0.013797 & merged & SimpleCNN\_2D & returns\\\hline
\end{tabular}
\caption{Forecast results for various sampling, aggregation types, models and tasks. Sorted by MAPE value}
\label{tab:table1}
\end{table}

\begin{table}[ht!]
\centering
\begin{tabular}{|l|r|r|}
\hline
\textbf{Model} & \textbf{Aggregation}  & \textbf{MAPE, \%} \\\hline
CNN2LSTM      & stacked  & 0.018244  \\\hline
SimpleCNN     & stacked & 0.103163  \\\hline
CNNModel\_2D  & merged  & 0.344536 \\\hline
SimpleCNN\_2D & merged  & 0.377812  \\\hline
\end{tabular}
\caption{Models mid-price forecasting average performance. Sorted by MAPE value}
\label{tab:table2}
\end{table}

\begin{table}[ht!]
\centering
\begin{tabular}{|l|r|}
\hline
\textbf{Aggregation} & \textbf{MAPE, \%} \\\hline
stacked & 0.060703 \\\hline
merged & 0.361174\\\hline
\end{tabular}
\caption{Aggregation method average performance.}
\label{tab:table3}
\end{table}

The Table~\ref{table:4features_horizon} represents experiments runs with samples constructed with 30 consecutive ($30w$ sampling type) LOB snapshots for various prediction horizons. 
The further the horizon is, the higher the error is. 
Minimum values of MAPE are highlighted in bold.
The error increases by 10 times when the horizon reaches $60s$. The result is clear, because the closer the target value, the higher correlation. We also observe that in most cases the predicting the price delta is more suitable for short-term forecasting, but returns is better for long-term forecasting.

We also performed experiments with the number of LOB a channel features increased to~8, incorporating volume global scaling and price bins difference as described in Section 3.1. The Table~\ref{table:8features_horizon} reflects the results. The \textit{BTC} and \textit{ETH} mid-price forecasting models benefit from additional futures in case of the $0.2s$ horizon. We cannot accurately conclude whether the features are beneficial or not, because we did not performed any hyperparameter tuning, but the error rate unexpectedly increased having jump at $0.2s$ horizon. 

\begin{figure}[ht!]
    \centering
    \includegraphics[width=1.0\linewidth]{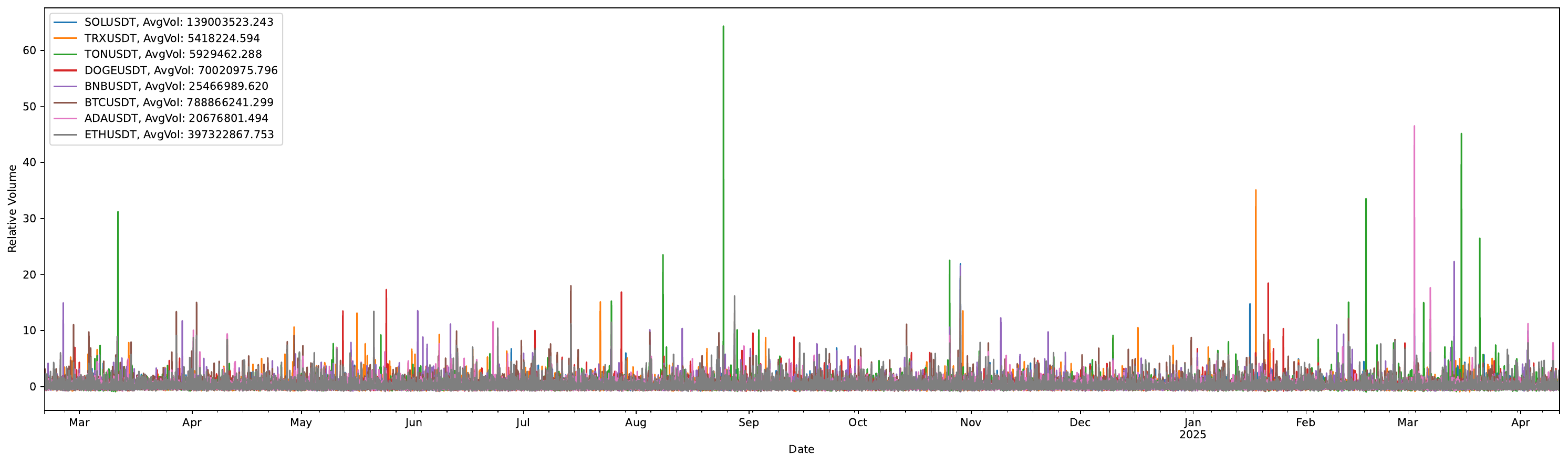}
    \caption{Volume relative changes, February 20, 2024 to April 4, 2025}
    \label{fig:vol_change}
\end{figure}

\begin{figure}[ht!]
    \centering
    \includegraphics[width=0.75\linewidth]{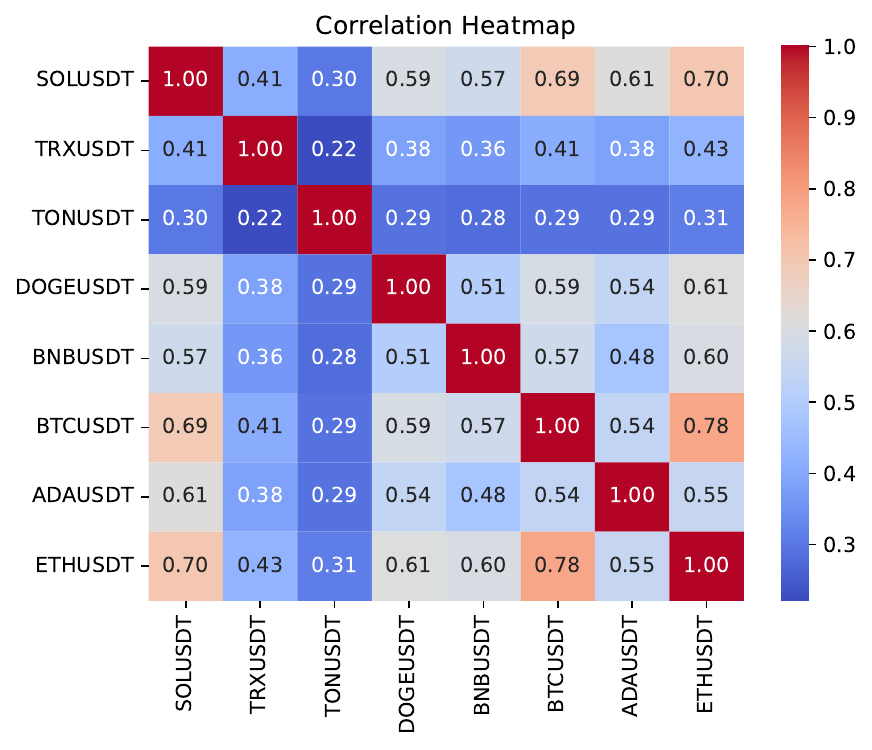}
    \caption{Volumes change correlation heat map, February 20, 2024 to April 4, 2025}\label{fig:vol_corr_heat_map}
\end{figure}

We also examined whether we can train a single model to process a group of assets. We performed experiments in three setups: single, mixed and one shot. In the single setting we train separate models for each asset. In the mixed case we aggregate assets LOB snapshots into a single dataset and train model to predict delta or returns on various assets, building a kind of general model. In the one shot setup we aggregate data over assets and train a model to predict mid-prices simultaneously. To perform training we first analyzed assets to find a grouping pattern. The Figure~\ref{fig:vol_change} represents assets volume relative changes. According to the volumes change correlation heat map in Figure~\ref{fig:vol_corr_heat_map} the only \textit{BTC} and \textit{ETH} futures have a strong correlation greater than $0.7$. Although \textit{SOL} is one of the most liquid cryptocurrencies and has greater correlation with BTC and ETH
than others, we found the correlation to be low.

Having found the pair (BTC, ETH), we first trained a model with mixed dataset: sample of both assets were used in the training, on test the model predicted mid-prices of the cryptocurrencies one after another. For the one-shot case, we first aggregated both assets LOB snapshots and stacked them to feed to the \textit{CNN2LSTM} model, doubling the number of the input image channels. The model predicted both features mid-prices simultaneously. For the experiments, we used historical interval of the length of 30 consequent snapshots with 4 and 8 feature columns for all 6 forecasting horizons. The experiment results are presented in the Table~\ref{table:btc_hor}. The \textit{MAPE} error is presented in bps; $4F$ and $8F$ postfixes denote the number of features columns included into the LOB image: 4 and 8 features respectfully. The one-shot setting helps to significantly reduce the error value approximately make it 4 times less in comparison with the single setting in case of the $0.2s$ and $1s$ forecasting horizons. One-shot and mixed training helps to reduce \textit{MAPE} at the task of returns forecasting at short forecasting horizons. It discovered 
that predicting returns turn to be a more robust on long horizons, while predicting deltas is better for short-term trend forecasting.

{\centering
\subsection{Trading Strategy}}

\begin{figure}
    \centering
    \includegraphics[width=0.7\linewidth]{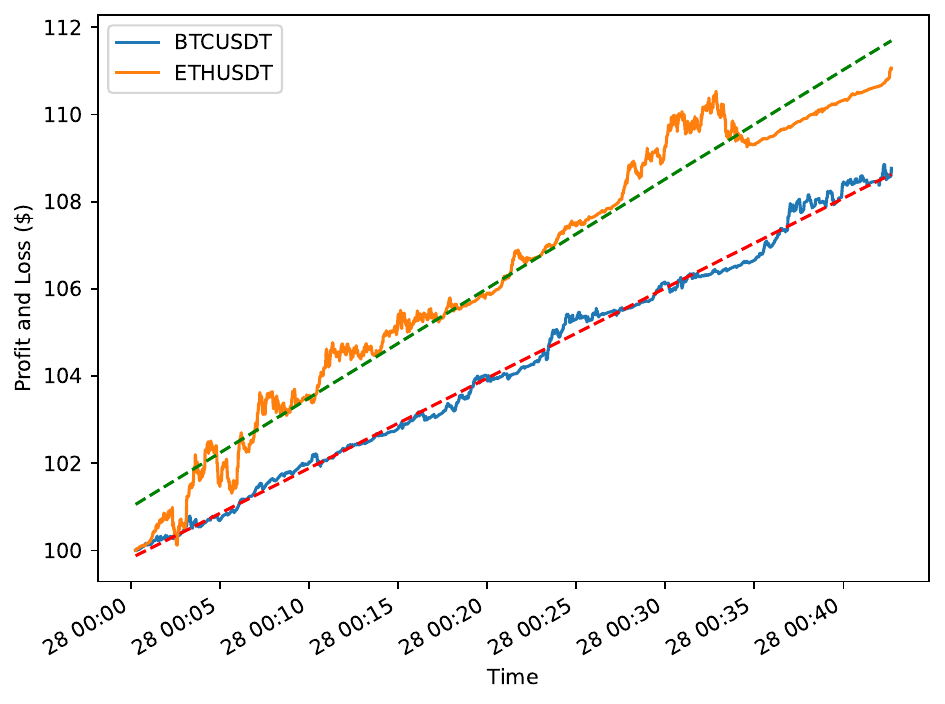}
    \caption{Profit and Loss curve for stacked method, aggregation interval 30~\textit{w} (in U.S. dollars).  The straight line plots the linear regressions fit. 
BTC: 
$PnL = 99.87 + 17.61 \cdot 10^{-4} \cdot \ln(time)$ and 
ETH: $PnL = 101.05 + 21.4\cdot 10^{-4}\cdot \ln(time)$
}
    \label{fig:trading_result}
\end{figure}

We used a single model, that was trained on BTC 
LOB data to execute trading of both BTC and ETH futures. The trading strategy base on CNN2LSTM model predictions with $30$ frames length sliding window results is presented in Figure~\ref{fig:trading_result}. We dashed green and red lines are linear approximation of BTC and ETH strategies performance. The strategy demonstrates strong performance, having spread earning slightly positive, while generating sufficient trading volume to earn on negative fees. In comparison with the model presented, the SimpleCNN\_2D with merged LOBs and price delta regression task had poor performance. The model tended to repeat the price trend, it had low trading volume and tendency to hold long position.

The Table~\ref{tab:avg_growth} describes the strategy characteristics. For profit and loss lines velocity means average growth rate, and slope value for regression lines, and is measured in USD bps. 

\begin{table}[ht!]
\centering
\begin{tabular}{|l|c|c|c|}
\hline
\textbf{Line} & \textbf{Velocity (bps)} & \textbf{Bias} & \textbf{Sharpe}\\\hline
BTCUSDT	& 17.64 & 100.00 & 7.89\\\hline
BTCUSDT Regression & 17.61 & 99.87 & --\\\hline
ETHUSDT	& 22.26 & 100.00 & 4.48\\\hline
ETHUSDT Regression & 21.40 & 101.05 & --\\\hline

\end{tabular}
\caption{Trading strategies average growth and their linear regression slope value.}
\label{tab:avg_growth}
\end{table}

{\centering
\section{Conclusion and Further Work}}

We showed that the proposed approach to treat LOB snapshots as a sequential image channels performs better than merging them into a single 2-dimensional image. 

Although the issue of choosing the optimal LOB sequence length remains open, the question of determining the appropriate target value is partially solved. Price delta is better for short-term prediction forecasting, but the mid-price relative growth value is better for long-term forecasting.

It is essential to notice that although min-max domain scaling offers better in-domain distribution information, it might lack cross domain correlation between ask and bid volume distributions. The correlation is important for decision making because it reflects the market state such as volume imbalance. That is why the issue of complementary features naturally arises. Proposed additional features: volume global scaling and price bins differences partially address the issue.

It is important to investigate the limitation issue of orders execution. In the current research, we have very optimistic strategy, considering that the minimal order value will definitely be executed. In fact, the real exchange is much complicated, there is a risk the order will not be executed because of orders queue etc.

Knowing the limit order book state and forecasting its evolution dynamic might help to build more robust to execution algorithmic strategies, it is essential to incorporate the designed LOB processing framework into middle frequency trading strategies.

We also see a good potential to integrate reinforcement learning into the proposed pipeline with ``financially`` grounded reward policies, that will make model training more intuitive and trustworthy for traders.

\noindent
\textbf{Funding}
\textit{
This work was supported by the grant of the state program of the ``Sirius`` Federal Territory ``Scientific and technological development of the ``Sirius`` Federal Territory`` (Agreement No.~18-03 data 10.09.2024).
}

\noindent
\textbf{Data availability} The used dataset is published and accessible on Kaggle platform: \\ \url{https://www.kaggle.com/datasets/kkhubiev/btc-lob/data} 

\begin{landscape}
\begin{table}[ht!]
    \centering
    \caption{Forecasting performance (MAPE, bps) for tickers with different horizons and target values, 4 LOB feathers}   \label{table:4features_horizon}
  \resizebox{1.5\textheight}{!}{
    \begin{tabular}{|l|rr|rr|rr|rr|rr|rr|}
        \hline
        \multirow{2}{*}{\centering{Ticker}} & 
        \multicolumn{2}{c|}{0.2s} &
        \multicolumn{2}{c|}{1s} &
        \multicolumn{2}{c|}{5s} &
        \multicolumn{2}{c|}{10s} &
        \multicolumn{2}{c|}{30s} &
        \multicolumn{2}{c|}{60s} \\
        \cline{2-13}
        & delta & returns 
        & delta & returns 
        & delta & returns 
        & delta & returns 
        & delta & returns 
        & delta & returns 
        \\\hline
        ADA
        & \textbf{4.0608} & \textbf{4.0684} 
        & 6.5488 & 6.5609 
        & 10.5594 & 10.5599 
        & 11.8042 & 10.5599 
        & 23.4401 & 11.0921 
        & 35.0604 & 34.9744 
        \\\hline
        BNB
        & \textbf{0.4473} & 3.5378 
        & 1.1215 & \textbf{3.0642} 
        & 2.6080 & 4.5389 
        & 3.4464 & 6.1253 
        & 8.0489 & 10.9732 
        & 12.5134 & 13.2562 
        \\\hline
        BTC
        & \textbf{0.4039} & \textbf{1.7114} 
        & 1.0030 & 1.9755 
        & 2.5335 & 3.4717 
        & 3.5881 & 4.4318 
        & 7.7604 & 8.0355 
        & 11.7311 & 11.7082 
        \\\hline
        DOGE
        & \textbf{5.5022} & \textbf{2.7826} 
        & 5.6041 & 3.6401 
        & 8.398 & 7.1593 
        & 11.3433 & 9.1489 
        & 19.1686 & 18.1120 
        & 28.3736 & 27.3628 
        \\\hline
        ETH
        & \textbf{0.8210} & \textbf{4.1362} 
        & 1.5031 & 4.6131 
        & 3.5291 & 5.6284 
        & 5.2785 & 6.5229 
        & 10.5565 & 11.0921 
        & 14.1682 & 14.5061 
        \\\hline
        GOAT
        & \textbf{4.0729} & \textbf{3.6337} 
        & 6.5817 & 5.9020 
        & 9.7854 & 10.2692 
        & 13.6341 & 13.3915 
        & 27.2481 & 27.0113 
        & 37.687 & 36.4944 
        \\\hline
        TON
        & \textbf{1.4299} & \textbf{3.3976} 
        & 2.7562 & 3.9514 
        & 6.001 & 7.1998 
        & 9.0914 & 9.3516 
        & 17.7552 & 18.1472 
        & 26.7795 & 29.1056 
        \\\hline
        TRX
        & \textbf{6.8002} & \textbf{1.5505} 
        & 10.4883 & 2.4671 
        & 12.7779 & 3.5136 
        & 8.9803 & 3.7547 
        & 12.2555 & 6.5367 
        & 18.6451 & 9.8755 
        \\\hline
        SOL
        & \textbf{0.7063} & \textbf{2.7658} 
        & 1.2677 & 4.1875 
        & 3.2531 & 4.4746 
        & 4.6380 & 6.2026 
        & 9.1552 & 10.247 
        & 13.8838 & 16.5093 
        \\\hline
    \end{tabular}
}
\end{table}    
\end{landscape}

\begin{landscape}
\begin{table}[ht!]
    \centering
   \caption{Forecasting performance (MAPE, bps) for tickers with different horizons and target values, 8 LOB feathers}
   \label{table:8features_horizon}
  \resizebox{1.5\textheight}{!}{
    \begin{tabular}{|l|rr|rr|rr|rr|rr|rr|}
        \hline
        \multirow{2}{*}{\centering{Ticker}} & 
        \multicolumn{2}{c|}{0.2s} &
        \multicolumn{2}{c|}{1s} &
        \multicolumn{2}{c|}{5s} &
        \multicolumn{2}{c|}{10s} &
        \multicolumn{2}{c|}{30s} &
        \multicolumn{2}{c|}{60s} \\
        \cline{2-13}
        & delta & returns 
        & delta & returns 
        & delta & returns 
        & delta & returns 
        & delta & returns 
        & delta & returns 
        \\\hline
        ADA
        & 14.7935 & 10.4293 
        & \textbf{8.0403} & \textbf{7.9663} 
        & 11.4527 & 10.6675 
        & 14.8065 & 13.9341 
        & 23.6611 & 23.9627 
        & 36.0946 & 36.2004 
        \\\hline
        BNB
        & \textbf{0.4390} & 7.7688 
        & 0.9780 & \textbf{5.3715} 
        & 2.7434 & 5.6927 
        & 3.9233 & 12.4767 
        & 8.0440 & 15.0792 
        & 11.6071 & 16.3937 
        \\\hline
        BTC
        & \textbf{0.3704} & 3.7822 
        & 0.9156 & \textbf{2.3593} 
        & 2.5390 & 4.0339 
        & 3.7001 & 4.6769 
        & 7.8176 & 8.4597 
        & 11.7454 & 12.6412 
        \\\hline
        DOGE
        & 32.7192 & \textbf{3.2928} 
        & \textbf{7.5245} & 6.0137 
        & 9.7734 & 7.9286 
        & 15.7924 & 10.5756 
        & 22.8965 & 19.8353 
        & 31.3757 & 29.5532 
        \\\hline
        ETH
        & \textbf{0.7764} & \textbf{3.9465} 
        & 1.5478 & 6.4999 
        & 3.6996 & 8.1708 
        & 5.7442 & 8.7679 
        & 10.6389 & 12.9193 
        & 14.3498 & 15.1891 
        \\\hline
        GOAT
        & \textbf{7.8885} & \textbf{6.4449} 
        & 8.2145 & 9.8294 
        & 11.4967 & 11.8752 
        & 16.6933 & 15.9896 
        & 28.9768 & 27.9951 
        & 40.7777 & 39.8277 
        \\\hline
        TON
        & \textbf{1.7084} & \textbf{5.5706} 
        & 2.7888 & 5.9366 
        & 6.4460 & 8.2944 
        & 9.6412 & 12.0253 
        & 17.7242 & 20.2976 
        & 25.7449 & 30.0969 
        \\\hline
        TRX
        & 23.4831 & 12.8756 
        & \textbf{19.3399} & \textbf{5.0191} 
        & 25.1937 & 6.1920 
        & 36.8325 & 7.3911 
        & 48.1647 & 8.4999 
        & 79.7617 & 11.5546 
        \\\hline
        SOL
        & \textbf{0.6788} & 4.7958 
        & 1.3607 & \textbf{4.7888} 
        & 3.4446 & 6.1424 
        & 5.0128 & 8.0607 
        & 9.0184 & 10.8922 
        & 13.1890 & 15.8166 
        \\\hline
    \end{tabular}
}
\end{table}
\end{landscape}

\begin{landscape}
\begin{table}[ht!]
\centering
   \caption{Model's forecasting performance (MAPE, bps) for BTC, ETH  with different horizons and target values, 4 and 8 LOB feathers}\label{table:btc_hor}
  \resizebox{1.5\textheight}{!}{
    \begin{tabular}{|l|rr|rr|rr|rr|rr|rr|}
        \hline
        \multicolumn{13}{|c|}{BTC}
        \\\hline
        \multirow{2}{*}{\centering{Model}} & 
        \multicolumn{2}{c|}{0.2s} &
        \multicolumn{2}{c|}{1s} &
        \multicolumn{2}{c|}{5s} &
        \multicolumn{2}{c|}{10s} &
        \multicolumn{2}{c|}{30s} &
        \multicolumn{2}{c|}{60s} \\
        \cline{2-13}
        & delta & returns 
        & delta & returns 
        & delta & returns 
        & delta & returns 
        & delta & returns 
        & delta & returns 
        \\\hline
        Single 4F
        & 0.4048 & 1.7114 
        & 1.0032 & 1.9755 
        & 2.5337 & 3.4717 
        & 3.5883 & 4.4318 
        & 7.7604 & 8.0355 
        & 11.7311 & 11.7080 
        \\\hline
        Mixed 4F
        & 0.4051 & 0.7939 
        & 1.004 & 1.4607 
        & \textbf{2.5331} & 2.8572 
        & \textbf{3.5875} & \textbf{3.7164} 
        & 7.7598 & 7.6370 
        & 11.7295 & 11.5760 
        \\\hline
        One Shot 4F
        & 0.4052 & 1.2113 
        & 1.0036 & 1.3535 
        & \textbf{2.5331} & 2.8329 
        & 3.5881 & 3.9512 
        & \textbf{7.7584} & 7.8507 
        & \textbf{11.7292} & 11.4855 
        \\\hline
        Single 8F
        & 0.3704 & 3.7822 
        & 0.9156 & 2.3593 
        & 2.5390 & 4.0339 
        & 3.7001 & 4.6769 
        & 7.8176 & 8.4597 
        & 11.7454 & 12.6412 
        \\\hline
        Mixed 8F
        & \textbf{0.3691} & 1.5147 
        & \textbf{0.9151} & 1.3259 
        & 2.5390 & 2.8935 
        & 3.6995 & 3.9481 
        & 7.8164 & \textbf{7.6074} 
        & 11.7447 & 11.7647 
        \\\hline
        One Shot 8F
        & 0.3723 & \textbf{0.4901} 
        & 0.9207 & \textbf{1.0849} 
        & 2.5390 & \textbf{2.6448} 
        & 3.7020 & 3.7716 
        & 7.8208 & 7.9061 
        & 11.748 & \textbf{11.4266} 
        \\\hline
        \multicolumn{13}{|c|}{ETH}
        \\\hline
        \multirow{2}{*}{\centering{Model}} & 
        \multicolumn{2}{c|}{0.2s} &
        \multicolumn{2}{c|}{1s} &
        \multicolumn{2}{c|}{5s} &
        \multicolumn{2}{c|}{10s} &
        \multicolumn{2}{c|}{30s} &
        \multicolumn{2}{c|}{60s} \\
        \cline{2-13}
        & delta & returns 
        & delta & returns 
        & delta & returns 
        & delta & returns 
        & delta & returns 
        & delta & returns 
        \\\hline
        Single 4F
        & 0.8237 & 4.1361 
        & \textbf{1.5045} & 4.6105 
        & 3.5275 & 5.6284 
        & 5.2810 & 6.5229 
        & \textbf{10.5570} & 11.0921 
        & \textbf{14.1681} & 14.5003 
        \\\hline
        Mixed 4F
        & 0.8223 & 2.5423 
        & 1.5048 & 2.9301 
        & \textbf{3.5184} & 5.1383 
        & \textbf{5.2632} & 6.4410 
        & 10.5649 & 11.1592 
        & 14.1869 & 14.7516 
        \\\hline
        One Shot 4F
        & 0.8136 & 1.1673 
        & 1.4964 & \textbf{1.7318} 
        & 3.5372 & 4.1682 
        & 5.3003 & \textbf{5.6292} 
        & 10.6494 & 10.5734 
        & 14.3053 & \textbf{13.3077} 
        \\\hline
        Single 8F
        & 0.7764 & 3.9465 
        & 1.5478 & 6.4999 
        & 3.6996 & 8.1708 
        & 5.7442 & 8.7679 
        & 10.6389 & 12.9193 
        & 14.3498 & 15.1891 
        \\\hline
        Mixed 8F
        & 0.7728 & 2.9434 
        & 1.5435 & 3.0960 
        & 3.7018 & 5.3449 
        & 5.7571 & 7.3277 
        & 10.6575 & 12.2598 
        & 14.3861 & 15.4388 
        \\\hline
        One Shot 8F
        & \textbf{0.7666} & \textbf{0.8241} 
        & 1.5430 & 1.8555 
        & 3.7874 & \textbf{3.8724} 
        & 5.8725 & 5.7762 
        & 10.8632 & \textbf{10.5564} 
        & 14.6696 & 13.8308 
        \\\hline
    \end{tabular}
    }
\end{table}
\end{landscape}

\bibliographystyle{unsrt}

\end{document}